\documentclass{article}
\usepackage{spconf,amsmath,amssymb,booktabs,graphicx,multirow,url,microtype,array}
\usepackage[table]{xcolor}
\usepackage[T1]{fontenc}
\usepackage[utf8]{inputenc}
\usepackage{balance,placeins,fix-cm,needspace}
\usepackage{cite}
\usepackage{hyperref}
\definecolor{groupA}{HTML}{6C7685}
\definecolor{groupB}{HTML}{2863B7}
\definecolor{groupC}{HTML}{21846A}
\definecolor{groupD}{HTML}{B74F60}
\definecolor{groupAt}{HTML}{F4F5F7}
\definecolor{groupBt}{HTML}{EFF5FD}
\definecolor{groupCt}{HTML}{EFF9F5}
\definecolor{groupDt}{HTML}{FDF1F3}

\newcommand{\mstd}[2]{#1\,$\pm #2$}
\newcommand{\bmstd}[2]{\textbf{#1}\,$\mathbf{\pm #2}$}

\title{DO WE NEED COMPLEX TOPOLOGY CONTROL? DISTINCT-PEER RANDOM ROUTING IMPROVES COST-EFFICIENCY IN SPARSE MULTI-AGENT DEBATE}

\name{
Boxuan Wang \qquad
Zhuoyun Li \qquad
Xiaowei Huang \qquad
Yi Dong\textsuperscript{\textdagger}
}

\address{School of Computer Science and Informatics, University of Liverpool, United Kingdom}

\begin{document}
\ninept
\raggedbottom
\maketitle

\begingroup
\renewcommand{\thefootnote}{\textdagger}
\footnotetext{Corresponding author: yi.dong@liverpool.ac.uk.}
\endgroup

\begin{abstract}
Multi-agent debate (MAD) has emerged as a promising paradigm for improving the reasoning accuracy of large language models (LLMs) through iterative peer interaction. Communication topology plays a central role in this process, motivating increasingly sophisticated mechanisms that learn, adapt, or dynamically reconfigure agent interactions to improve accuracy or reasoning reliability. Meanwhile, prior studies suggest that much simpler sparse communication can already achieve competitive performance at substantially lower cost. In this work, we take a closer look at sparse MAD and ask whether complex topology control is actually necessary to improve collective reasoning. We find that a simple random-without-replacement routing policy, which lets each agent debate with two distinct and newly sampled peers at every round, provides a surprisingly strong baseline and consistently improves the accuracy-cost trade-off of sparse MAD. Building on this observation, we further study deliberation stopping and show that lightweight stopping can substantially reduce inference cost while preserving competitive accuracy. Our results suggest that sophisticated topology control such as learned topology adaption should be evaluated against strong simple routing and stopping baselines before its additional complexity is justified.
\end{abstract}

\section{Introduction}
Multi-agent debate (MAD) is widely regarded as a paradigm that can improve reasoning quality of large language models (LLMs) 
through proposal, critique, and revision among agents
\cite{du2024debate,liang2024divergent,pmlr-v235-smit24a,pmlr-v235-khan24a}.
Its communication topology determines whose intermediate solutions each
agent can inspect, shaping both opportunities for correction and the spread
of redundant or erroneous information. Prior work shows that sparse
communication can match or sometimes exceed fully connected debate in
answer accuracy using fewer tokens
\cite{li2024sparse,zeng-etal-2025-s2}.

\begin{figure}[t]
\centering
\includegraphics[width=\columnwidth]{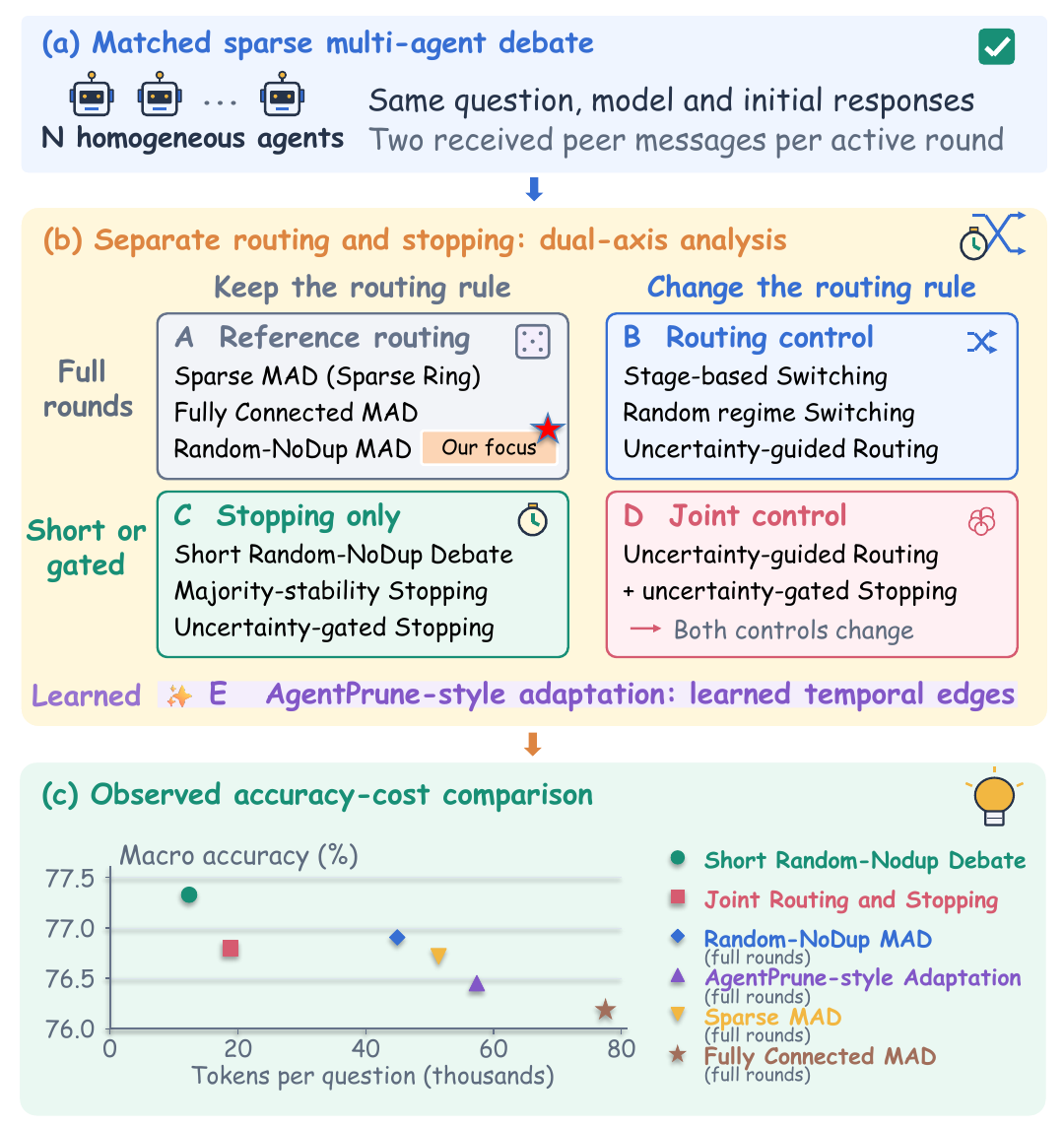}
\caption{An overview of the study. (a) Homogeneous sparse debate as the starting point. (b) The dual-axis routing
and stopping evaluation framework. (c) Observed accuracy-cost comparison.}
\label{fig:design}
\end{figure}

Recent topology-control methods learn sparse masks, generate task-adaptive
graphs, select interaction structures, or use graph diffusion
\cite{zhang2025agentprune,zhang2025gdesigner,leong2025amas,jiang2026gtd,dong2023decentralised}.
The motivation is intuitive: better peer selection may retain useful
information while avoiding redundant or misleading messages. Yet a simple
neighbor-connected topology already offers competitive answer accuracy at
lower token cost than dense debate \cite{li2024sparse}. Analyses of
homogeneous debate also find that repeated interaction need not outperform
majority voting over initial responses \cite{zhu2026demystifying}. We
therefore ask: \emph{does additional topology control improve accuracy or
reduce token cost beyond simple sparse routing and short debate?}

We study this question through a \textbf{two-axis analysis framework}
(Fig.~\ref{fig:design}). Routing determines \emph{who communicates}, while
stopping determines \emph{how long communication continues}. We compare
reference topologies, schedule-based and uncertainty-guided routing,
stopping-only policies, and their combination. A matched-protocol
temporal-edge adaptation of AgentPrune provides a learned comparator
\cite{zhang2025agentprune}. This design separates the effect of choosing
peers from the effect of reducing the number of rounds.

We focus on \textbf{Random-NoDup}, which samples distinct peers in each
round and allows later revisits. We measure final-answer accuracy and total
input plus output tokens per question. At full rounds, Random-NoDup has
higher observed macro accuracy than Ring at similar token cost. Its short
version has the highest observed macro accuracy and lowest mean token use
among the tested methods.

In this paper, we primarily ask two research questions: \textbf{RQ1}, how competitive are simple routing baselines compared with structured, adaptive, and learned topology
control? \textbf{RQ2}, does additional stopping control further improve the
accuracy--cost trade-off? Our contributions are: \textbf{(1)} we introduce Random-NoDup as a simple
distinct-peer sparse routing baseline; \textbf{(2)} we propose a two-axis
evaluation framework for MAD, and empirically showing that Random-NoDup is
highly competitive and stopping further improves its efficiency.

\section{Related Work}
\textbf{Sparse multi-agent debate.} Early MAD lets multiple model instances propose and revise solutions through repeated peer exchange \cite{du2024debate,chen2024reconcile}. Li et al. systematically vary communication connectivity and show that sparse MAD can match or outperform fully connected debate while reducing computation \cite{li2024sparse}. Later work makes sparse communication more selective: CortexDebate constructs a sparse debating graph in which agents interact with peers estimated to be useful \cite{sun2025cortex}, while AgentPrune removes redundant edges from the spatial--temporal message-passing graph \cite{zhang2025agentprune}. Shen et al. study information propagation directly, finding that moderate sparsity can suppress error diffusion while preserving beneficial information flow, and build EIB-Learner from this observation \cite{shen2025propagation}. Complementarily, recent analysis of homogeneous MAD shows that additional debate is not automatically useful and that lightweight diversity or confidence interventions can outperform vanilla deliberation \cite{zhu2026demystifying,cui-etal-2026-free}. These results motivate sparse and economical communication, but leave open how strong a simple random sparse baseline can be when peer collisions and deliberation length are controlled.

\noindent\textbf{Topology control in MAD.}
A parallel line treats the communication graph itself as an optimization
variable. GPTSwarm optimizes agent computational graphs
\cite{zhuge2024gptswarm,ICLR2025_66a026c0}; G-Designer generates task-adaptive
topologies \cite{zhang2025gdesigner,10.1609/aaai.v40i28.39481}; and AMAS dynamically selects
graph structures \cite{leong2025amas}. More recent work further
learns communication structure through guided diffusion
\cite{jiang2026gtd,doi:10.3233/FAIA251326} or reinforcement learning
\cite{cang2026graphgrpo}. Small-world priors provide another structured
alternative for balancing local and long-range communication
\cite{wang2025smallworld}. Our focus is complementary: rather than designing a
more expressive router, we ask whether topology control provides a clear
advantage over strong simple routing under a matched sparse MAD protocol. We
therefore compare against Sparse MAD and an AgentPrune-style learned
temporal-edge adaptation, while evaluating the effect of stopping.

\nocite{fragileflow,scope,pathmark,chen2026promptsleavebehavioralfingerprints,10980439,azulay2026jailbreaking,wang-etal-2026-chain,wang2026diveambiguityainspiredmultiagents}

\begin{table*}[!t] 
\centering 

\caption{Five-benchmark results. Accuracy (\%) and tokens (k/question) are mean $\pm$ std over five seeds; macro values are mean-only. Bold marks the best observed mean for each benchmark, not significance. \textbf{\textcolor{red}{Red}} numbers denote the best macro accuracy or token cost across methods.} 

\label{tab:online} 

\begingroup 
\fontsize{9}{9.6}\selectfont 
\renewcommand{\arraystretch}{0.92} 
\setlength{\tabcolsep}{2.0pt} 

\begin{tabular*}{\textwidth}{@{\extracolsep{\fill}}l>{\raggedright\arraybackslash}p{.265\textwidth}l*{6}{c}@{}} 

\toprule 

ID & Method & Metric 
& \shortstack{ARC-\\Challenge} 
& \shortstack{ScienceQA-\\text} 
& GSM8K 
& MMLU-Pro 
& \shortstack{GPQA-\\Diamond} 
& Macro \\ 

\midrule 

\rowcolor{groupAt} 
\multicolumn{9}{@{}l}{\strut\textbf{A\quad Reference routing and prior sparse topology}} \\ 

\multirow{2}{*}{A1} 
& \multirow{2}{=}{Ring / Sparse MAD} 
& Acc.$\uparrow$ 
& \mstd{90.67}{1.15} 
& \bmstd{96.00}{0.58} 
& \mstd{90.67}{3.06} 
& \mstd{63.33}{3.06} 
& \mstd{40.67}{1.15} 
& 76.27 \\ 

& & Tok.$\downarrow$ 
& \mstd{35.99}{0.35} 
& \mstd{33.32}{0.11} 
& \mstd{67.76}{0.10} 
& \mstd{64.74}{0.14} 
& \mstd{84.43}{0.58} 
& 57.25 \\ 

\multirow{2}{*}{A2} 
& \multirow{2}{=}{Small-world-inspired} 
& Acc.$\uparrow$ 
& \mstd{90.67}{1.15} 
& \mstd{95.33}{1.15} 
& \mstd{92.67}{1.15} 
& \mstd{64.67}{2.31} 
& \mstd{42.00}{0.00} 
& 77.07 \\ 

& & Tok.$\downarrow$ 
& \mstd{35.90}{0.09} 
& \mstd{33.15}{0.20} 
& \mstd{68.00}{0.10} 
& \mstd{65.10}{0.40} 
& \mstd{85.46}{0.51} 
& 57.52 \\ 

\multirow{2}{*}{A3} 
& \multirow{2}{=}{Random-NoDup, full deliberation} 
& Acc.$\uparrow$ 
& \mstd{91.33}{1.15} 
& \mstd{95.67}{1.15} 
& \mstd{90.00}{3.46} 
& \mstd{65.33}{3.06} 
& \mstd{42.00}{3.46} 
& 76.87 \\ 

& & Tok.$\downarrow$ 
& \mstd{35.69}{0.04} 
& \mstd{33.17}{0.11} 
& \mstd{67.95}{0.18} 
& \mstd{65.04}{0.28} 
& \mstd{85.04}{0.50} 
& 57.38 \\ 

\rowcolor{groupBt} 
\multicolumn{9}{@{}l}{\strut\textbf{B\quad Routing control with full deliberation}} \\ 

\multirow{2}{*}{B1} 
& \multirow{2}{=}{Stage-based topology switching} 
& Acc.$\uparrow$ 
& \mstd{90.00}{2.00} 
& \bmstd{96.00}{2.00} 
& \mstd{88.00}{2.00} 
& \mstd{63.33}{1.15} 
& \mstd{41.33}{4.16} 
& 75.73 \\ 

& & Tok.$\downarrow$ 
& \mstd{35.79}{0.15} 
& \mstd{33.29}{0.18} 
& \mstd{67.70}{0.15} 
& \mstd{65.10}{0.20} 
& \mstd{85.07}{1.13} 
& 57.39 \\ 

\multirow{2}{*}{B2} 
& \multirow{2}{=}{Random topology switching} 
& Acc.$\uparrow$ 
& \mstd{91.33}{1.15} 
& \mstd{94.00}{2.00} 
& \mstd{91.33}{1.15} 
& \mstd{63.33}{4.16} 
& \mstd{39.33}{4.16} 
& 75.86 \\ 

& & Tok.$\downarrow$ 
& \mstd{35.70}{0.14} 
& \mstd{33.27}{0.09} 
& \mstd{68.00}{0.10} 
& \mstd{65.10}{0.60} 
& \mstd{85.43}{0.78} 
& 57.50 \\ 

\multirow{2}{*}{B3} 
& \multirow{2}{=}{Uncertainty-guided routing} 
& Acc.$\uparrow$ 
& \mstd{90.67}{1.15} 
& \mstd{94.00}{1.39} 
& \bmstd{93.33}{1.15} 
& \mstd{64.00}{2.00} 
& \mstd{40.67}{3.06} 
& 76.53 \\ 

& & Tok.$\downarrow$ 
& \mstd{35.78}{0.23} 
& \mstd{33.23}{0.11} 
& \mstd{67.80}{0.20} 
& \mstd{65.00}{0.20} 
& \mstd{84.81}{1.02} 
& 57.32 \\ 

\rowcolor{groupCt} 
\multicolumn{9}{@{}l}{\strut\textbf{C\quad Random-NoDup with stopping control}} \\ 

\multirow{2}{*}{C1} 
& \multirow{2}{=}{\textbf{Random-NoDup, fixed round 3}} 
& Acc.$\uparrow$ 
& \mstd{92.00}{3.46} 
& \mstd{94.00}{2.00} 
& \mstd{92.00}{0.00} 
& \mstd{65.33}{1.15} 
& \mstd{43.33}{2.31} 
& \textcolor{red}{\textbf{77.33}} \\ 

& & Tok.$\downarrow$ 
& \bmstd{8.05}{0.03} 
& \bmstd{7.41}{0.03} 
& \bmstd{13.11}{0.02} 
& \bmstd{14.01}{0.07} 
& \bmstd{19.20}{0.13} 
& \textcolor{red}{\textbf{12.36}} \\ 

\multirow{2}{*}{C2} 
& \multirow{2}{=}{Majority-stability stopping} 
& Acc.$\uparrow$ 
& \mstd{92.00}{2.00} 
& \mstd{92.67}{1.15} 
& \mstd{90.67}{1.15} 
& \mstd{62.00}{0.00} 
& \mstd{43.33}{4.16} 
& 76.13 \\ 

& & Tok.$\downarrow$ 
& \mstd{8.13}{0.03} 
& \mstd{7.42}{0.01} 
& \mstd{13.80}{0.40} 
& \mstd{16.20}{0.60} 
& \mstd{23.55}{0.75} 
& 13.82 \\ 

\multirow{2}{*}{C3} 
& \multirow{2}{=}{Uncertainty-gated stopping} 
& Acc.$\uparrow$ 
& \bmstd{93.33}{1.15} 
& \mstd{94.00}{0.79} 
& \mstd{92.00}{0.00} 
& \mstd{62.67}{5.03} 
& \bmstd{44.00}{5.29} 
& 77.20 \\ 

& & Tok.$\downarrow$ 
& \mstd{9.72}{0.64} 
& \mstd{7.75}{0.05} 
& \mstd{16.59}{1.82} 
& \mstd{20.66}{0.20} 
& \mstd{38.27}{2.35} 
& 18.60 \\ 

\rowcolor{groupDt} 
\multicolumn{9}{@{}l}{\strut\textbf{D\quad Joint routing and stopping control}} \\ 

\multirow{2}{*}{D1} 
& \multirow{2}{=}{Uncertainty-guided routing + stopping} 
& Acc.$\uparrow$ 
& \mstd{92.67}{1.15} 
& \mstd{93.33}{1.15} 
& \mstd{92.00}{2.00} 
& \mstd{64.00}{0.00} 
& \mstd{42.00}{4.00} 
& 76.80 \\ 

& & Tok.$\downarrow$ 
& \mstd{9.66}{0.82} 
& \mstd{7.74}{0.07} 
& \mstd{17.60}{0.90} 
& \mstd{19.30}{1.80} 
& \mstd{39.90}{7.69} 
& 18.84 \\ 

\rowcolor{gray!10} 
\multicolumn{9}{@{}l}{\strut\textbf{E\quad Learned topology adaptation}} \\ 

\multirow{2}{*}{E1} 
& \multirow{2}{=}{AgentPrune-style temporal-edge adaptation} 
& Acc.$\uparrow$ 
& \mstd{91.33}{1.15} 
& \mstd{94.00}{2.00} 
& \mstd{89.33}{3.06} 
& \bmstd{66.67}{4.62} 
& \mstd{40.00}{5.29} 
& 76.27 \\ 

& & Tok.$\downarrow$ 
& \mstd{35.75}{0.20} 
& \mstd{33.39}{0.12} 
& \mstd{67.88}{0.20} 
& \mstd{65.21}{0.03} 
& \mstd{85.70}{1.13} 
& 57.59 \\ 

\bottomrule 

\end{tabular*} 

\endgroup 

\end{table*}

\section{preliminaries and Methods}
\label{sec:method}
\subsection{Sparse debate and reference topologies}
\begin{figure}[t]
\centering
\includegraphics[width=\columnwidth]{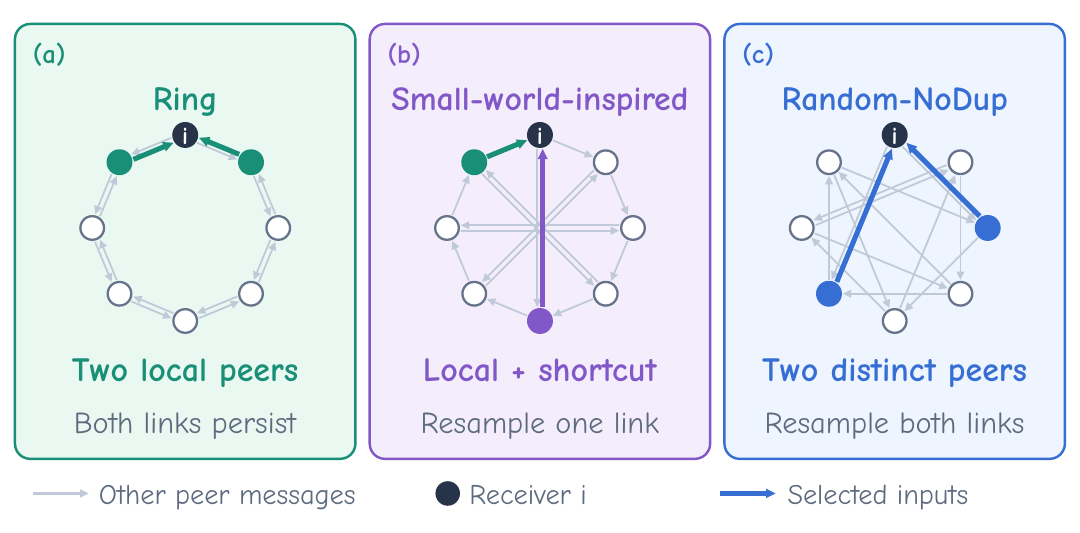}
\caption{Three reference routing rules with matched received degree. The
example graphs share node positions; colored arrows highlight two inputs to
receiver $i$. (a) Ring keeps both local peers. (b) Small-world-inspired routing
keeps one local peer and resamples one non-local shortcut. (c) Random-NoDup
resamples two distinct peers. Previously contacted peers remain eligible.}
\label{fig:topology_adv}
\end{figure}

We use $n$ homogeneous agents $\mathcal V$ and at most $T$ answer rounds.
Agents answer independently at $t=1$; later updates read $k$ peers' responses
from the previous round. Let $a_i^{(t)}$ denote the extracted answer of agent
$i$ at round $t$, and let $\mathcal Y_t$ be the set of distinct answers
produced by all agents at that round. For each $y\in\mathcal Y_t$, we define
the empirical answer distribution
\begin{equation}
 \begin{aligned}
 p_t(y)&=\frac{1}{n}\sum_i\mathbf1[a_i^{(t)}=y],
 & m_t&\in\arg\max_{y\in\mathcal Y_t} p_t(y),\\
 C_t&=p_t(m_t),
 & H_t&=-\sum_{y\in\mathcal Y_t} p_t(y)\log p_t(y).
 \end{aligned}
 \label{eq:answer_state}
\end{equation}
Here, $p_t(y)$ is the fraction of agents predicting answer $y$, $m_t$ is the
modal (majority) answer, $C_t$ is the fraction of agents supporting that
answer, and $H_t$ measures disagreement among agent answers. In the following, we provide the descriptions and illustrations (as shown in Fig.~\ref{fig:topology_adv}) of our reference baselines:

\noindent\textbf{Ring / Sparse MAD.}
Each receiver keeps the same local peers across rounds
(Fig.~\ref{fig:topology_adv}(a)), yielding stable local information flow.
This instantiates the neighbor-connected topology under our protocol
\cite{li2024sparse}.

\noindent\textbf{Small-world-inspired routing.}
Each receiver combines persistent local communication with a resampled
non-local shortcut (Fig.~\ref{fig:topology_adv}(b)). The local connection
preserves continuity across rounds, while the shortcut exposes the receiver to
a changing source of information.

\noindent\textbf{Random-NoDup routing.}
Each receiver draws $k$ distinct peers at every active round
(Fig.~\ref{fig:topology_adv}(c)):
\begin{equation}
 \mathcal N_i^{(t)}\sim\operatorname{Unif}
 \bigl\{S\subseteq\mathcal V\setminus\{i\}:|S|=k\bigr\}.
 \label{eq:nodup}
\end{equation}
Earlier peers remain eligible. The graph varies across rounds, but the rule
is state-agnostic. Distinct peers need not provide independent information.

\subsection{Routing and stopping controls}
Stage-based switching follows a preset schedule, while random switching selects
a routing regime independently of agent states. Uncertainty-guided routing uses
\begin{equation}
U_t=\frac{H_t}{2\log n}
+\frac{1}{2n}\sum_{i=1}^{n}
\mathbf1[a_i^{(t)}\ne a_i^{(t-1)}], \qquad t\ge2,
\label{eq:u}
\end{equation}
where $H_t$ is answer entropy, $n$ is the number of agents, and
$a_i^{(t)}$ is agent $i$'s answer at round $t$. Thresholds
$\theta_{\mathrm{low}}<\theta_{\mathrm{high}}$ map low, intermediate, and high
$U_t$ to Ring, small-world-inspired, and Random-NoDup routing. Thus $U_t$
captures disagreement and answer changes rather than correctness.

Short Random-NoDup returns the majority answer $m_h$ at a fixed horizon $h<T$.
Adaptive stopping uses
\begin{equation}
\begin{aligned}
g_t&=\mathbf1[m_t=m_{t-1}]
     \mathbf1[U_t<\theta_{\mathrm{stop}}],\\
\tau&=\min\!\left(\{t:\prod_{j=0}^{p-1}g_{t-j}=1\}\cup\{T\}\right),
\end{aligned}
\label{eq:stopping}
\end{equation}
where $\theta_{\mathrm{stop}}$ is the uncertainty threshold, $p$ is the required
number of stable transitions, and $\tau$ is the stopping round. The parameters are selected using held-out calibration sets.
Removing the uncertainty condition gives majority-stability stopping, while
joint control combines uncertainty-guided routing with the same stopping rule.

\subsection{Learned temporal-edge adaptation}
We also include an AgentPrune-style learned topology baseline
\cite{zhang2025agentprune}. AgentPrune assigns learnable probabilities to
candidate communication edges, updates them from task rewards, and prunes
low-value edges to obtain a sparse topology. For a matched comparison, we
adapt its learned edge-selection mechanism to our sparse MAD setting using
disjoint calibration questions, while keeping the agents, prompts,
synchronous updates, received degree, and aggregation rule unchanged. This
provides a learned topology-control baseline under the same sparse debate
protocol as the other methods.


\setlength{\textfloatsep}{6pt plus 1pt minus 1pt}
\setlength{\dbltextfloatsep}{6pt plus 1pt minus 1pt}
\makeatletter
\let\savedsubsection\subsection
\renewcommand{\subsection}{\@startsection{subsection}{2}{\z@}
  {-1.8ex plus -.3ex minus -.2ex}{.7ex plus .1ex}{\normalfont\large\bfseries}}
\makeatother
\clubpenalties 2 10000 0
\widowpenalties 2 10000 0
\section{Experiments}
\subsection{Experimental Setup}
\noindent\textbf{Homogeneous agents and benchmarks.}
We study homogeneous MAD, where all eight agents in a debate use the same
LLM. The main experiments use GPT-4o-mini on ARC-Challenge
\cite{clark2018arc}, ScienceQA-text \cite{lu2022scienceqa}, GSM8K
\cite{cobbe2021gsm8k}, MMLU-Pro \cite{wang2024mmlupro}, and GPQA-Diamond
\cite{rein2024gpqa}. ScienceQA-text contains text-only questions.
across models.

\noindent\textbf{Debate protocol.}
Each debate contains eight agents, and each agent receives messages from two
peers per communication round. Round 1 consists of independent answers,
followed by synchronous debate updates. Full debate runs for 14 rounds, while
Short Random-NoDup stops at round 3 after two communication updates. All
methods use the same generation settings (e.g., temperature and prompts) within each matched comparison.
Results are reported as mean $\pm$ standard deviation over five independent
runs with random seeds.

\noindent\textbf{Routing and stopping controls.}
We compare the reference topologies with stage-based, random, and
uncertainty-guided routing, as well as fixed and adaptive stopping. Stage-based
routing switches after round 5. For uncertainty-based control, we use
$\theta_{\mathrm{low}}=\theta_{\mathrm{stop}}=1/3$,
$\theta_{\mathrm{high}}=2/3$, and patience $p=2$. The learned-edge baseline
uses separate calibration questions. We evaluate each method by final-answer
accuracy and inference token cost, with macro results averaged equally across
tasks.

\subsection{RQ1: Does Topology Control Improve Routing?}
\label{sec:routing_results}
We test whether topology control improves accuracy beyond simple routing
by comparing reference rules and controllers at the same degree and full
horizon. Table~\ref{tab:online} reports accuracy and token cost.

\noindent\textbf{Reference routing.}
Ring ties for the highest ScienceQA-text accuracy, while
small-world-inspired routing has the highest macro accuracy among the
three reference rules. Random-NoDup exceeds Ring in mean accuracy on
ARC-Challenge, MMLU-Pro, and GPQA-Diamond, and in macro accuracy, with
similar token use. Figure~\ref{fig:topology_adv} shows their key difference:
fixed local peers, one refreshed shortcut, or two resampled distinct peers.
No single pattern gives the highest accuracy on every task.

\noindent\textbf{Additional routing control.}
Full-round Random-NoDup has higher macro accuracy than all three routing
controllers and the learned-edge adaptation. Uncertainty-guided routing
leads on GSM8K, and the learned adaptation leads on MMLU-Pro, but these
gains do not carry across tasks. Full-round token costs remain close,
and learned edges require calibration. Additional control can therefore
help individual tasks without improving the overall trade-off.

\noindent\textbf{Overall comparison.}
Across the matched full-round comparisons, Random-NoDup remains competitive
with structured, adaptive, and learned routing while requiring no state
estimation or topology optimization. Results show that additional topology control does not
provide a consistent accuracy--cost advantage over this baseline. We
next examine whether stopping offers a clearer efficiency gain.

\begin{table}[!t]
\centering
\caption{Cross-model accuracy (\%) and tokens (k/question): mean above $\pm$ SD. Bold: best mean per model--task pair. Round 3 is a same-trajectory cutoff; other runs use full rounds.}
\label{tab:cross_model}
\begingroup
\fontsize{9}{10}\selectfont
\setlength{\tabcolsep}{1.1pt}
\renewcommand{\arraystretch}{1.00}
\newcommand{\cmstd}[2]{\shortstack[c]{#1\\[-1.0pt]$\pm$#2}}
\newcommand{\cbmstd}[2]{\shortstack[c]{\textbf{#1}\\[-1.0pt]$\boldsymbol{\pm}$\textbf{#2}}}
\begin{tabular*}{\columnwidth}{@{\extracolsep{\fill}}l*{6}{c}@{}}
\toprule
& \multicolumn{2}{c}{\shortstack{ARC-\\Challenge}} & \multicolumn{2}{c}{GSM8K} & \multicolumn{2}{c}{\shortstack{GPQA-\\Diamond}} \\
\cmidrule(lr){2-3}\cmidrule(lr){4-5}\cmidrule(l){6-7}
Method & Acc.$\uparrow$ & Tok.$\downarrow$ & Acc.$\uparrow$ & Tok.$\downarrow$ & Acc.$\uparrow$ & Tok.$\downarrow$ \\
\midrule
\rowcolor{groupBt}\multicolumn{7}{@{}l@{}}{\strut\textbf{GPT-4o-mini}} \\
\shortstack[l]{Ring /\\[-1pt]Sparse MAD} & \cmstd{90.67}{1.15} & \cmstd{35.99}{0.35} & \cmstd{90.67}{3.06} & \cmstd{67.76}{0.10} & \cmstd{40.67}{1.15} & \cmstd{84.43}{0.58} \\[0pt]
\shortstack[l]{Random-NoDup\\[-1pt](full rounds)} & \cmstd{91.33}{1.15} & \cmstd{35.69}{0.04} & \cmstd{90.00}{3.46} & \cmstd{67.95}{0.18} & \cmstd{42.00}{3.46} & \cmstd{85.04}{0.50} \\[0pt]
\rowcolor{groupCt}\shortstack[l]{Random-NoDup\\[-1pt](round 3)} & \cbmstd{92.67}{2.31} & \cbmstd{8.04}{0.00} & \cmstd{90.00}{0.00} & \cbmstd{13.12}{0.03} & \cbmstd{48.67}{5.03} & \cbmstd{19.26}{0.12} \\[0pt]
\shortstack[l]{Uncertainty-guided\\[-1pt]routing} & \cmstd{90.67}{1.15} & \cmstd{35.78}{0.23} & \cbmstd{93.33}{1.15} & \cmstd{67.78}{0.02} & \cmstd{40.67}{3.06} & \cmstd{84.81}{1.02} \\[0pt]
\addlinespace[1pt]
\rowcolor{groupBt}\multicolumn{7}{@{}l@{}}{\strut\textbf{GPT-4.1-mini}} \\
\shortstack[l]{Ring /\\[-1pt]Sparse MAD} & \cbmstd{94.67}{1.15} & \cmstd{35.30}{0.16} & \cmstd{94.00}{2.00} & \cmstd{53.38}{0.32} & \cmstd{62.67}{1.15} & \cmstd{96.69}{0.14} \\[0pt]
\shortstack[l]{Random-NoDup\\[-1pt](full rounds)} & \cbmstd{94.67}{1.15} & \cmstd{35.31}{0.16} & \cmstd{94.00}{0.00} & \cmstd{53.50}{0.72} & \cmstd{66.00}{2.00} & \cmstd{98.08}{0.34} \\[0pt]
\rowcolor{groupCt}\shortstack[l]{Random-NoDup\\[-1pt](round 3)} & \cmstd{94.00}{2.00} & \cbmstd{8.18}{0.01} & \cbmstd{94.67}{1.15} & \cbmstd{10.56}{0.02} & \cbmstd{66.67}{3.06} & \cbmstd{23.53}{0.15} \\[0pt]
\shortstack[l]{Uncertainty-guided\\[-1pt]routing} & \cmstd{92.67}{1.15} & \cmstd{35.43}{0.12} & \cmstd{94.00}{2.00} & \cmstd{53.65}{0.66} & \cmstd{63.33}{1.15} & \cmstd{97.58}{0.44} \\[0pt]
\addlinespace[1pt]
\rowcolor{groupBt}\multicolumn{7}{@{}l@{}}{\strut\textbf{GPT-4.1}} \\
\shortstack[l]{Ring /\\[-1pt]Sparse MAD} & \cmstd{92.67}{1.15} & \cmstd{27.89}{0.02} & \cmstd{96.67}{1.15} & \cmstd{40.00}{0.10} & \cbmstd{62.67}{1.15} & \cmstd{54.13}{0.48} \\[0pt]
\shortstack[l]{Random-NoDup\\[-1pt](full rounds)} & \cmstd{93.33}{1.15} & \cmstd{27.88}{0.02} & \cmstd{96.00}{2.00} & \cmstd{40.13}{0.03} & \cbmstd{62.67}{1.15} & \cmstd{53.89}{0.26} \\[0pt]
\rowcolor{groupCt}\shortstack[l]{Random-NoDup\\[-1pt](round 3)} & \cbmstd{94.67}{1.15} & \cbmstd{7.02}{0.01} & \cmstd{91.33}{1.15} & \cbmstd{9.06}{0.04} & \cmstd{60.67}{1.15} & \cbmstd{17.38}{0.16} \\[0pt]
\shortstack[l]{Uncertainty-guided\\[-1pt]routing} & \cmstd{92.00}{0.00} & \cmstd{27.88}{0.01} & \cbmstd{97.33}{1.15} & \cmstd{40.19}{0.18} & \cbmstd{62.67}{3.06} & \cmstd{54.10}{0.34} \\[0pt]
\addlinespace[1pt]
\rowcolor{groupBt}\multicolumn{7}{@{}l@{}}{\strut\textbf{GPT-5.6 Luna}} \\
\shortstack[l]{Ring /\\[-1pt]Sparse MAD} & \cmstd{84.67}{1.15} & \cmstd{26.86}{0.01} & \cmstd{98.00}{0.00} & \cmstd{26.80}{0.01} & \cmstd{54.00}{3.46} & \cmstd{46.77}{0.02} \\[0pt]
\shortstack[l]{Random-NoDup\\[-1pt](full rounds)} & \cmstd{84.67}{1.15} & \cmstd{26.86}{0.01} & \cmstd{98.67}{1.15} & \cmstd{26.80}{0.01} & \cmstd{54.67}{4.16} & \cmstd{46.78}{0.00} \\[0pt]
\rowcolor{groupCt}\shortstack[l]{Random-NoDup\\[-1pt](round 3)} & \cbmstd{88.00}{0.00} & \cbmstd{5.93}{0.00} & \cbmstd{99.33}{1.15} & \cbmstd{6.42}{0.01} & \cbmstd{58.67}{2.31} & \cbmstd{10.89}{0.01} \\[0pt]
\shortstack[l]{Uncertainty-guided\\[-1pt]routing} & \cmstd{84.00}{0.00} & \cmstd{26.86}{0.01} & \cmstd{97.33}{2.31} & \cmstd{26.79}{0.01} & \cmstd{52.00}{2.00} & \cmstd{46.77}{0.03} \\[0pt]
\bottomrule
\end{tabular*}
\endgroup

\end{table}

\subsection{RQ2: Does Stopping Improve the Trade-off?}
\label{sec:stopping_results}
We next keep Random-NoDup routing fixed and compare full rounds, a short
fixed cutoff, and two answer-based stopping gates. This experiment evaluates whether implementing an additional stopping criterion can increase accuracy while reducing token consumption.

\noindent\textbf{Fixed and adaptive stopping.}
Short Random-NoDup achieves the best observed macro accuracy while using the
fewest tokens (Table~\ref{tab:online}). Adaptive stopping can improve some
individual tasks, but at higher cost, and adding routing control does not
improve the overall trade-off. Paired results consistently favor the short
setting in token cost, while accuracy differences are generally small.

\begin{figure}[!t]
\centering
\includegraphics[width=.88\columnwidth]{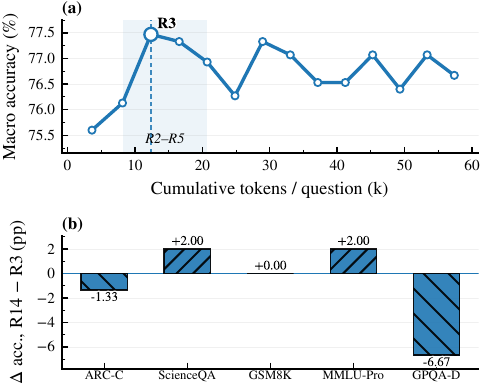}
\caption{Random-NoDup horizon analysis. (a) Accuracy versus token cost for
cutoffs of the same trajectories; the shaded region covers rounds 2--5.
(b) Per-task accuracy change from round 3 to round 14.}
\label{fig:horizon}
\end{figure}

\noindent\textbf{Why use round 3?}
Figure~\ref{fig:horizon}(a) tests cutoffs of complete trajectories. Macro
accuracy is close to its observed peak during rounds 2--5, while later
rounds add tokens. Round 3 includes two communication updates, and its
paired accuracy intervals against rounds 2, 4, and 5 include zero. It is
an early reference, not a uniquely optimal cutoff. These prefixes are
separate from the independent short runs in Table~\ref{tab:online}.

\noindent\textbf{Overall accuracy--cost trade-off.}
Across the stopping variants, the main benefit of short debate is improved
efficiency rather than a uniform accuracy gain. As shown in
Table~\ref{tab:online}, the short Random-NoDup setting achieves accuracy
comparable to the longer and adaptive alternatives while using substantially
fewer tokens. Figure~\ref{fig:horizon} further shows that most of the useful
performance is already reached in the early rounds, whereas additional debate
continues to increase cost without a consistent accuracy improvement. Thus,
for RQ2, short debate provides a stronger accuracy--cost trade-off by reducing
inference cost while preserving competitive accuracy.

\subsection{Ablations on Cross-model Performance}
\label{sec:cross_model_results}
We repeat four key comparisons across four model variants to check whether
the findings depend on the primary model. Table~\ref{tab:cross_model}
compares Ring, full and short Random-NoDup, and uncertainty-guided routing
on the same three tasks. Short results use the round-3 prefix of each
matched full Random-NoDup trajectory.

\noindent\textbf{Routing across models.}
At full rounds, neither Ring nor uncertainty-guided routing establishes a
paired accuracy advantage over Random-NoDup in any of the twelve
model--task comparisons. The intervals also do not establish consistent
superiority for Random-NoDup. Thus, simple random routing remains a
competitive reference across the tested variants, not only in the primary
model.

\noindent\textbf{Stopping across models.}
Short Random-NoDup uses the fewest tokens for every model--task pair and
has the highest macro accuracy for three of the four models. However, one
model benefits from longer debate on GSM8K, where stopping at round 3
reduces accuracy. The cost savings transfer more consistently than accuracy
preservation. Overall, a short debate paradigm with distinct-peer routing constitutes a robust baseline that generalizes across the evaluated models.

\section{Conclusion}
In this paper, we study sparse multi-agent debate through a two-axis framework that
separates peer routing from stopping. The comparisons cover reference
topologies, switching policies, uncertainty-guided control, joint control,
and a matched-protocol AgentPrune-style temporal-edge adaptation.
Experimental results show that the proposed distinct-peer random routing method, Random-NoDup, provides competitive accuracy without learned or state-aware
routing, while short debate further improves the token savings.

Cross-model results support this distinction: additional routing control
shows no detected accuracy advantage over Random-NoDup in the tested
comparisons, whereas stopping consistently reduces token use. Its accuracy
effect still depends on the model and task. Our findings indicate that advanced topology control methods, including learned topology adaptation and switching methods should first be benchmarked against the proposed distinct-peer random routing baselines before their added complexity can be warranted.

\begin{center}\bf ACKNOWLEDGEMENTS\end{center}
This work is partially funded by the European Union (under grant agreement ID 101212818). Views and opinions expressed are however those of the author(s) only and do not necessarily reflect those of the European Union or European Health and Digital Executive Agency (HADEA). Neither the European Union nor the granting authority can be held responsible for them. 
This work is partially supported by Innovate UK through AI-PASSPORT under Grant 10126404. This work was awarded a grant by the AI Security Institute (AISI) via the Alignment Project
(Rare-Event Estimation in Large Language Models via Subset Simulation) and funded by EPSRC.
Yi's contribution is partially supported through the Royal Society international exchanges programme and in part by the Engineering and Physical Sciences Research Council, through funding from RAi UK [EP/Y009800/1].

\begin{center}
\bfseries COMPLIANCE WITH ETHICAL STANDARDS
\end{center}
This work involves no human participants or newly collected human-subject data. All experiments rely on public benchmarks and model inference conducted under applicable terms of use.

\bibliographystyle{IEEEbib}
\bibliography{references}

@inproceedings{du2024debate,
  author    = {Y. Du and S. Li and A. Torralba and J. B. Tenenbaum and I. Mordatch},
  title     = {Improving factuality and reasoning in language models through multiagent debate},
  booktitle = {Proc. ICML},
  year      = {2024}
}

@inproceedings{li2024sparse,
  author    = {Y. Li and Y. Du and J. Zhang and L. Hou and P. Grabowski and Y. Li and others},
  title     = {Improving Multi-Agent Debate with Sparse Communication Topology},
  booktitle = {Findings ACL: EMNLP},
  pages     = {7281--7294},
  year      = {2024}
}

@inproceedings{sun2025cortex,
  author    = {Y. Sun and Z. Zhao and S. Wan and C. Gong},
  title     = {{CortexDebate}: Debating Sparsely and Equally for Multi-Agent Debate},
  booktitle = {Findings ACL},
  pages     = {9503--9523},
  year      = {2025}
}

@inproceedings{zhang2025agentprune,
  author    = {G. Zhang and Y. Yue and Z. Li and S. Yun and G. Wan and K. Wang and others},
  title     = {Cut the Crap: An Economical Communication Pipeline for {LLM}-based Multi-Agent Systems},
  booktitle = {Proc. ICLR},
  year      = {2025}
}

@inproceedings{shen2025propagation,
  author    = {X. Shen and Y. Liu and Y. Dai and Y. Wang and R. Miao and Y. Tan and others},
  title     = {Understanding the Information Propagation Effects of Communication Topologies in {LLM}-based Multi-Agent Systems},
  booktitle = {Proc. EMNLP},
  pages     = {12347--12361},
  year      = {2025}
}

@inproceedings{zhu2026demystifying,
  author    = {X. Zhu and C. Zhang and Y. Chi and T. Stafford and N. Collier and A. Vlachos},
  title     = {Demystifying Multi-Agent Debate: The Role of Confidence and Diversity},
  booktitle = {Findings ACL},
  pages     = {33909--33930},
  year      = {2026}
}

@inproceedings{zhuge2024gptswarm,
  author    = {M. Zhuge and W. Wang and L. Kirsch and F. Faccio and D. Khizbullin and J. Schmidhuber},
  title     = {{GPTSwarm}: Language Agents as Optimizable Graphs},
  booktitle = {Proc. ICML},
  pages     = {62743--62767},
  year      = {2024}
}

@inproceedings{zhang2025gdesigner,
  author    = {G. Zhang and Y. Yue and X. Sun and G. Wan and M. Yu and J. Fang and others},
  title     = {G-Designer: Architecting Multi-agent Communication Topologies via Graph Neural Networks},
  booktitle = {Proc. ICML},
  year      = {2025}
}

@inproceedings{leong2025amas,
  author    = {H. Y. Leong and Y. Li and Y. Wu and W. Ouyang and W. Zhu and J. Gao and others},
  title     = {{AMAS}: Adaptively Determining Communication Topology for {LLM}-based Multi-agent System},
  booktitle = {Proc. EMNLP Industry Track},
  pages     = {2061--2070},
  year      = {2025}
}

@inproceedings{jiang2026gtd,
  author    = {E. H. Jiang and L. Li and F. Wan and X. Liang and S. Yin and Y. Wu and others},
  title     = {Dynamic Generation of Multi {LLM} Agents Communication Topologies with Graph Diffusion Models},
  booktitle = {Proc. ACL},
  pages     = {38042--38060},
  year      = {2026}
}

@inproceedings{cang2026graphgrpo,
  author    = {Y. Cang and X. Zhang and E. Zhao and Z. Ji and Y. Liu and Y. He and others},
  title     = {Graph-{GRPO}: Stabilizing Multi-Agent Topology Learning via Group Relative Policy Optimization},
  booktitle = {Findings ACL},
  pages     = {20222--20231},
  year      = {2026}
}

@misc{wang2025smallworld,
  author = {B. Wang and Z. Li and X. Huang and Y. Dong},
  title  = {Rethinking Multi-Agent Intelligence Through the Lens of Small-World Networks},
  note   = {arXiv:2512.18094},
  year   = {2025}
}

@misc{clark2018arc,
  author = {P. Clark and I. Cowhey and O. Etzioni and T. Khot and A. Sabharwal and C. Schoenick and others},
  title  = {Think you have Solved Question Answering? Try {ARC}, the {AI2} Reasoning Challenge},
  note   = {arXiv:1803.05457},
  year   = {2018}
}

@misc{cobbe2021gsm8k,
  author = {K. Cobbe and V. Kosaraju and M. Bavarian and M. Chen and H. Jun and L. Kaiser and others},
  title  = {Training Verifiers to Solve Math Word Problems},
  note   = {arXiv:2110.14168},
  year   = {2021}
}

@inproceedings{lu2022scienceqa,
  author    = {P. Lu and S. Mishra and T. Xia and L. Qiu and K.-W. Chang and S.-C. Zhu and others},
  title     = {Learn to Explain: Multimodal Reasoning via Thought Chains for Science Question Answering},
  booktitle = {NeurIPS},
  year      = {2022}
}

@inproceedings{wang2024mmlupro,
  author    = {Y. Wang and X. Ma and G. Zhang and Y. Ni and A. Chandra and S. Guo and others},
  title     = {{MMLU}-Pro: A More Robust and Challenging Multi-Task Language Understanding Benchmark},
  booktitle = {NeurIPS Datasets and Benchmarks Track},
  year      = {2024}
}

@inproceedings{rein2024gpqa,
  author    = {D. Rein and B. L. Hou and A. C. Stickland and J. Petty and R. Y. Pang and J. Dirani and others},
  title     = {{GPQA}: A Graduate-Level Google-Proof Q\&A Benchmark},
  booktitle = {Proc. COLM},
  year      = {2024}
}

@inproceedings{dong2023decentralised,
  author    = {Y. Dong and Z. Li and X. Zhao and Z. Ding and X. Huang},
  title     = {Decentralised and Cooperative Control of Multi-Robot Systems through Distributed Optimisation},
  booktitle = {Proc. AAMAS},
  pages     = {1421--1429},
  year      = {2023}
}

@inproceedings{chen2024reconcile,
  author    = {J. Chen and S. Saha and M. Bansal},
  title     = {{ReConcile}: Round-Table Conference Improves Reasoning via Consensus among Diverse {LLM}s},
  booktitle = {Proc. ACL},
  pages     = {7066--7085},
  year      = {2024}
}

@inproceedings{liang2024divergent,
  author    = {T. Liang and Z. He and W. Jiao and X. Wang and Y. Wang and R. Wang and others},
  title     = {Encouraging Divergent Thinking in Large Language Models through Multi-Agent Debate},
  booktitle = {Proc. EMNLP},
  pages     = {17889--17904},
  year      = {2024}
}

@inproceedings{pmlr-v235-smit24a,
  author    = {A. P. Smit and N. Grinsztajn and P. Duckworth and T. D. Barrett and A. Pretorius},
  title     = {Should we be going {MAD}? {A} Look at Multi-Agent Debate Strategies for {LLM}s},
  booktitle = {Proc. ICML},
  pages     = {45883--45905},
  year      = {2024}
}

@inproceedings{zeng-etal-2025-s2,
  author    = {Y. Zeng and W. Huang and L. Jiang and T. Liu and X. Jin and C. T. Tiana and others},
  title     = {S$^2$-{MAD}: Breaking the Token Barrier to Enhance Multi-Agent Debate Efficiency},
  booktitle = {Proc. NAACL},
  pages     = {9393--9408},
  year      = {2025}
}

@inproceedings{pmlr-v235-khan24a,
  author    = {A. Khan and J. Hughes and D. Valentine and L. Ruis and K. Sachan and A. Radhakrishnan and others},
  title     = {Debating with More Persuasive {LLM}s Leads to More Truthful Answers},
  booktitle = {Proc. ICML},
  pages     = {23662--23733},
  year      = {2024}
}

@inproceedings{ICLR2025_66a026c0,
  author    = {C. Qian and Z. Xie and Y. Wang and W. Liu and K. Zhu and H. Xia and others},
  title     = {Scaling Large Language Model-based Multi-Agent Collaboration},
  booktitle = {Proc. ICLR},
  pages     = {41488--41505},
  year      = {2025}
}

@inproceedings{10.1609/aaai.v40i28.39481,
  author    = {S. Li and Y. Liu and Q. Wen and C. Zhang and S. Pan},
  title     = {Assemble Your Crew: Automatic Multi-Agent Communication Topology Design via Autoregressive Graph Generation},
  booktitle = {Proc. AAAI},
  year      = {2026},
  doi       = {10.1609/aaai.v40i28.39481}
}

@inproceedings{doi:10.3233/FAIA251326,
  author    = {B. Li and Z. Zhao and D.-H. Lee and G. Wang},
  title     = {Adaptive Graph Pruning for Multi-Agent Communication},
  booktitle = {Proc. ECAI},
  pages     = {4305--4312},
  year      = {2025},
  doi       = {10.3233/FAIA251326}
}

@inproceedings{cui-etal-2026-free,
  author    = {Y. Cui and H. Fu and H. Zhang and L. Wang and C. Zuo},
  title     = {Free-{MAD}: Consensus-Free Multi-Agent Debate},
  booktitle = {Findings ACL},
  pages     = {31977--31997},
  year      = {2026},
  doi       = {10.18653/v1/2026.findings-acl.1600}
}

@misc{fragileflow,
  author   = {Z. Li and B. Wang and J. Hu and X. Huang and Y. Dong},
  title    = {FragileFlow: Spectral Control of Correct-but-Fragile Predictions for Foundation Model Robustness},
  year     = {2026},
  note     = {arXiv:2605.08896}
}

@misc{scope,
  author   = {Z. Li and B. Wang and C. Wu and X. Huang and Y. Dong},
  title    = {SCOPE: Sequential Conformal Probing for Reliable OOD Rejection in LLM Services},
  year     = {2026},
  note     = {arXiv:2606.21255}
}

@misc{pathmark,
  author   = {Y. Gao and others},
  title    = {PathMark: Protecting Intellectual Property of Mixture-of-Expert LLMs via Path Watermarks},
  year     = {2026},
  note     = {arXiv:2607.03688}
}

@misc{chen2026promptsleavebehavioralfingerprints,
      title={Do System Prompts Leave Behavioral Fingerprints? A Large-Scale Empirical Study of Clone Detection via Output Similarity}, 
      author={Linghan Chen and Yudong Gao and Jiyao Wang and Kaiyan Ji and Honglong Chen},
      year={2026},
      eprint={2608.24461},
      archivePrefix={arXiv},
      primaryClass={cs.CR},
      url={https://arxiv.org/abs/2608.24461}, 
}

@ARTICLE{10980439,
  author={Yu, Jimiao and others},
  journal={IEEE Transactions on Multimedia}, 
  title={Black-Box Adversarial Defense Based on Image Decomposition and Reconstruction}, 
  year={2025},
  volume={27},
  number={},
  pages={5909-5921},
  doi={10.1109/TMM.2025.3565987}}

@inproceedings{
azulay2026jailbreaking,
title={Jailbreaking Vision-Language Models Through the Visual Modality},
author={Aharon Azulay and Jan Dubi{\'n}ski and Zhuoyun Li and Atharv Mittal and Yossi Gandelsman},
booktitle={Proc. ICML},
year={2026},
url={https://openreview.net/forum?id=4VuMV9beC1}
}

@inproceedings{wang-etal-2026-chain,
  author    = {B. Wang and Z. Li and X. Huang and X. Huang and Y. Dong},
  title     = {Chain-of-Thought as a Lens: Evaluating Structured Reasoning Alignment between Human Preferences and Large Language Models},
  booktitle = {Proc. ACL},
  pages     = {39514--39530},
  year      = {2026},
  doi       = {10.18653/v1/2026.acl-long.1834}
}

@misc{wang2026diveambiguityainspiredmultiagents,
  author = {B. Wang and Z. Li and X. Huang and Y. Dong},
  title  = {Dive into Ambiguity: A*-Inspired Multi-Agents Commonsense Obfuscation Attack on {LLM} Prompts},
  note   = {arXiv:2606.01441},
  year   = {2026}
}

\end{document}